\pdfoutput=1

\documentclass[twocolumn,11pt]{article}

\usepackage{arxiv_template}

\usepackage[T1]{fontenc}
\usepackage[utf8]{inputenc}
\usepackage{hyperref}       
\usepackage{url}            
\usepackage{booktabs}       
\usepackage{amsfonts}       
\usepackage{nicefrac}       
\usepackage{microtype}      
\usepackage{fancyhdr}       
\usepackage{graphicx}       

\usepackage{float}
\usepackage{bm}
\usepackage{amssymb}
\usepackage{multirow}
\usepackage{amsmath}
\usepackage[numbers]{natbib}
\usepackage{caption} 

\graphicspath{{img/}}

\DeclareMathOperator*{\argmax}{arg\,max}
\title{LSG Attention: Extrapolation of pretrained \\Transformers to long sequences}

\author{Charles Condevaux \\
  CHROME \\ University of Nîmes, France \\
  \texttt{charles.condevaux@unimes.fr} \\\And
  Sébastien Harispe \\
  EuroMov Digital Health in Motion,\\
  Univ Montpellier, IMT Mines Ales, France \\
  \texttt{sebastien.harispe@mines-ales.fr} \\}

\begin{document}
\maketitle

\begin{abstract}
Transformer models achieve state-of-the-art performance on a wide range of NLP tasks. They however suffer from a prohibitive limitation due to the self-attention mechanism, inducing $O(n^2)$ complexity with regard to sequence length. 
To answer this limitation we introduce the LSG architecture which relies on Local, Sparse and Global attention. 
We show that LSG attention is fast, efficient and competitive in classification and summarization tasks on long documents. Interestingly, it can also be used to adapt existing pretrained models to efficiently extrapolate to longer sequences with no additional training. Along with the introduction of the LSG attention mechanism, we propose tools to train new models and adapt existing ones based on this mechanism.
\end{abstract}

\section{Introduction}

Transformer models \cite{Vaswani2017} are nowadays state-of-the-art in numerous domains, and in particular in NLP where they are used in general language models, and to successfully tackle several specific tasks such as document summarization, machine translation and speech processing to cite a few \cite{Devlin2018,Raffel2020}. The cornerstone of Transformer models is the Attention mechanism used to iteratively build complex context-dependent representations of sequence elements, e.g. tokens, by dynamically aggregating prior representations of these elements. Using self-attention, a popular Attention flavour, this is made by computing full attention scores defining how each prior element representation will contribute to building the new representation of an element. Considering a sequence of $n$ elements, the computation of the attention scores is therefore of complexity $O(n^2)$ which is prohibitive when large sequences have to be processed. Furthermore, in the current context where a large number of models based on full attention have been trained on various datasets and tasks, we are also interested in extrapolating those models to longer sequences by simply substituting full attention by new attention mechanisms post training. Common pretrained models (e.g BERT, RoBERTa) are indeed known to underperform when extrapolated to sequences of length exceeding the 512 tokens considered during training. This is due to the nature of the attention mechanism which largely impacts extrapolation capabilities: full attention usually fails to extrapolate, even considering post hoc adaptations, e.g. adding constants in the score matrix \citep{Press2021}, using a relative positional embedding \citep{Shaw2018} or duplicating the positional embedding \citep{Beltagy2020}. Defining new attention mechanisms that can efficiently substitute full attention in pretrained models that are not originally capable of handling long sequences would avoid the costs induced by training large language models from scratch.

\noindent The main contributions of this paper are:

1. LSG (Local Sparse Global) attention, an efficient $O(n)$ approach to approximate self-attention for processing long sequences, is introduced.\footnote{\href{https://huggingface.co/ccdv}{https://huggingface.co/ccdv}}

2. Results demonstrating that LSG is fast, efficient and competitive on classification and summarization tasks applied to long documents are presented. It is also shown that LSG can adapt and extrapolate existing pretrained models not based on LSG, with minimal to no additional training.

3. A procedure and companion tools are proposed to convert various existing models and checkpoints (BERT, RoBERTa, DistilBERT, BART) from HuggingFace to their LSG variant.\footnote{\href{https://github.com/ccdv-ai/convert_checkpoint_to_lsg}{https://github.com/ccdv-ai/convert\_checkpoint\_to\_lsg}}

Compared to several contributions aiming at reducing the complexity of self-attention introduced hereafter, a specific focus is given in our work on the extrapolation of existing Transformer models, i.e. reuse, to longer sequences.

\section{Related works}

Several contributions have been devoted to the optimization of the Attention mechanism. 
Four categories of approaches can be distinguished in the literature: (i) recurrent models such as Transformers-XL \cite{Dai2019} and Compressive Transformers \cite{Rae2019} which maintain a memory of past activation at each layer to preserve long-range contextual information; (ii) models based on factorization or kernels aiming at compressing attention score matrices, such as Linformer \cite{Wang2020} or Performer \cite{Choromanski2021}; (iii) models based on clustering such as Reformer \cite{Kitaev2020} that dynamically define eligible attention patterns (i.e. where attention may be made); and (iv) models based on fixed or adaptative attention patterns, e.g. Longformer \cite{Beltagy2020} or Big Bird \cite{Zaheer2020}.

Recurrent approaches iteratively process the sequence by maintaining a memory to enable long-range dependencies. They generally suffer limitations induced by specific, slow, and difficult to implement forward and back propagation procedures. Alternatively, one of the main line of study for reducing the complexity of Attention is thus to perform sparsity by limiting the number of elements on which new representations will be based, i.e. reducing the number of elements with non-null attention scores. This approach is motivated by the observation of global or data-dependent positional patterns of non-null attention scores depending on the task \cite{Child2019}. The sparsity of attention scores in the traditional Attention mechanism is indeed documented in the literature. It has for instance been shown that in practice, full attention tends to overweight close elements in average, in particular for MLM, machine translation, and seq-to-seq tasks in general \cite{Clark2019}. Moreover, according to analyses on the use of multi-head full attention on specific tasks, e.g. machine translation, numerous heads learn similar simple patterns \cite{Raganato2020}. Such redundant patterns may be hardcoded implementing fixed-positional patterns, eventually in a task-dependent manner.

Two main approaches are discussed in the literature for implementing sparsity: fixed or adaptative patterns based on whether attention scores are computed considering (1) predefined fixed elements based on their location in the sequence, or (2) elements selected from a given procedure. 
As an example, \cite{Wu2019} have shown that fixed $O(n)$ convolutions can perform competitively on machine translation. Longformer proposes an alternative $O(n)$ approach based on sliding and global patterns \cite{Beltagy2020}. In the context of image, audio, and text processing, \cite{Child2019} propose sparse Transformer, an $O(n\sqrt{n})$ model based on sparse factorization of the attention matrix relying on specific 2D factorized attention schemes. Those approaches however prevent the use of task-dependent dynamic patterns. Considering adaptative patterns, \cite{Wu2019} also introduced dynamic convolutions as an $O(n)$ complexity substitute to self-attention. Kernels defining the importance of context elements are specified at inference time rather than fixed after training. Another example is Reformer \cite{Kitaev2020}, an $O(n ~log~n)$ approach based on locality-sensitive hashing (LSH) based on random projections. 

In a transverse manner, several authors, explicitly or implicitly motivated by the compositional nature of language have studied structured approaches in which subsequences (i.e. blocks) are processed independently and then aggregated. This aims at implementing a local or global dynamic memory for considering close to long-range dependencies. \cite{Britz2017} introduce a blockwise approach to reduce the quadratic complexity induced by large sequences in encoder-decoder architectures. \cite{Chiu2017} propose a chunkwise attention in which attention is performed in a blockwise manner adaptively splitting the sequence into small chunks over which soft attention is computed. This idea is also used in Transformer-XL \cite{Dai2019}. \cite{Shen2018} propose a masked block self-attention mechanism in which the entire sequence is divided into blocks, to further 1) apply self-attention intra-block for modeling local contexts, to further 2) apply self-attention inter-block for capturing long-range dependencies.
Such an approach enables implementing some forms of connectivity between all positions over several steps without being restricted by full attention limitations. This can also be achieved by factorization techniques, e.g. \cite{Child2019}.
More recently authors have proposed global attention mechanisms encoding information related to blocks on which attention is based \cite{Ainslie2020,Zhang2019b,Guo2019}.


This paper presents the LSG (Local, Sparse and Global) attention based on block local attention to capture local context, sparse attention to capture extended
context and global attention to improve information flow.
Contrary to prior work mostly focusing on defining new models, the proposed LSG Attention mechanism is model agnostic and aims at facilitating adapting existing (pretrained) models for them to be used on long sequences.

\section{LSG Attention}\label{section:lsg}

\begin{figure}[ht]
    \centering
    \includegraphics[width=\linewidth]{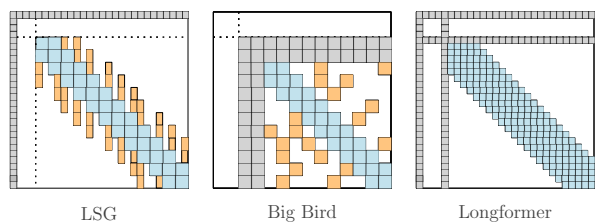}
    \caption{Attention patterns}
    \label{fig:attention}
\end{figure}

LSG attention relies on two main points. It is assumed that locally, a token needs to capture low level information thus dense attention is prefered. On the other hand, as the context grows, higher level information is sufficient. This translates into the need for connections to a limited number of tokens following specific selection and computation rules.
The LSG approach relies on 3 components: block local attention to capture local context, sparse attention to capture extended context and global attention to improve information flow. A comparison to Big Bird and Longformer attention patterns is shown in Figure \ref{fig:attention}.

\subsection{Local Attention}

Longformer depends on a fixed length sliding window to perform local attention. However this approach is difficult to optimize and must rely on a custom CUDA kernel to be computationally efficient. To improve overall training and inference speed, we take advantage of a block-based process similar to Big Bird. The sequence is split into $n_b$ non-overlapping chunks of size $b_t$. For a given block, each token attends to the tokens inside the block, as well as to those in the previous and next blocks. In this configuration, the local attention window is asymmetrical since a token can connect up to $2\times b_t - 1$ tokens on the left or on the right.

\subsection{Sparse Attention}

Sparse connections are used to expand the local context by selecting an additional set of tokens following a set of rules. These tokens can be directly selected based on a specific metric or using some computation such as a pooling method. In the proposed approach, each attention head can process different sparse tokens independently. Sparse attention also relies on a block structure where the sparse selection is done inside each block. Five alternative criteria can be used in LSG.

\begin{figure}[ht]
    \centering
    \includegraphics[width=\linewidth]{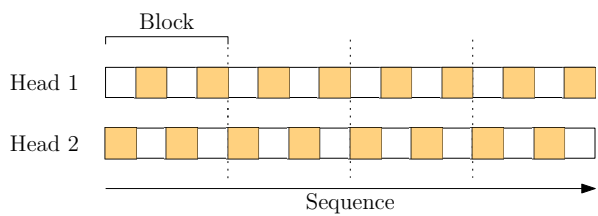}
    \caption{Head-wise strided selection with a stride of 2.}
    \label{fig:sparse_stride}
\end{figure}

\paragraph{Head-wise strided} Inspired by the Hepos model \citep{Huang2021}, a fixed selection pattern is defined. Each attention head will attend to a set of tokens following a specific stride defined as the sparsify factor $f$. Figure \ref{fig:sparse_stride} shows the selection pattern.

\paragraph{Head-wise block strided} This selection pattern is similar to the previous one but selects consecutive tokens instead. Figure \ref{fig:sparse_stride} shows the selection pattern.
\begin{figure}[ht]
    \centering
    \includegraphics[width=\linewidth]{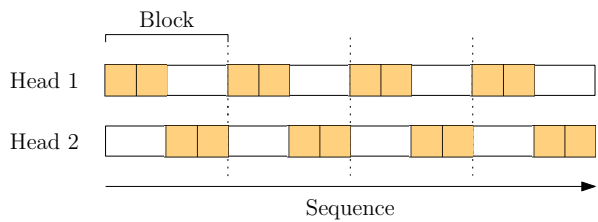}
    \caption{Block strided selection with a stride of 2.}
    \label{fig:sparse_blockstride}
\end{figure}

\paragraph{Average pooling} A simple way to reduce sequence length. After chunking the sequence into blocks, sparse tokens are computed using average pooling. For a block of size $b_t$ and a sparsify factor $f$, we pool inside each block with a window of $f$ and a stride of $f$ to produce $b_t/f$ tokens. 

\paragraph{Max norm} The objective of a norm-based approach is to select tokens that are most likely highly weighted in the score matrix. Finding those keys efficiently is difficult in practice so we use a simple and deterministic metric. For a query and a key $\bm{q},\bm{k} \in \mathbb{R}^d$, we can write:
$$\bm{q}\bm{k}^\top = \cos(\theta) \|\bm{q}\| \|\bm{k}\|$$
In this situation $cos(\theta)$ sign is unknown. However, if it is positive and $\|\bm{k}\|$ is high, the key will likely dominate the softmax regardless of the query. After chunking the sequence into blocks, we select inside each block and each head $b_t/f$ tokens with the highest key norm.

\paragraph{LSH Clustering} This approach is a non deterministic one since it relies on the LSH algorithm \citep{Andoni2015}. For each block, $b_t/f$ clusters are built using a single round LSH. To get $c = b_t/f$ hashes and for an input $\bm{x} \in \mathbb{R}^d$, a random matrix $\bm{R} \in \mathbb{R}^{d \times c/2}$ is generated, such that 
$$h(\bm{x}) = \argmax([\bm{x}\bm{R}; - \bm{x}\bm{R}])$$
with $[\bm{a}; \bm{b}]$ the concatenation of two vectors. Using the key matrix as input, each token inside the block gets a cluster index from $h(\bm{x})$. Tokens inside a cluster are averaged. 

\begin{figure}[ht]
    \centering
    \includegraphics[width=\linewidth]{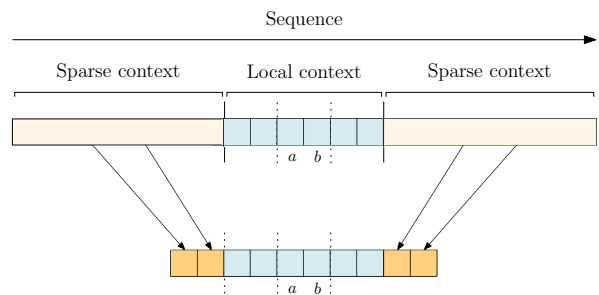}
    \caption{Local and sparse contexts with a block size of 2 and a sparsity factor of 4. Queries $a$ and $b$ will attend to 6 local keys and 4 sparse keys.}
    \label{fig:construction}
\end{figure}

\paragraph{Computation}
To reduce the computational cost, the attention pattern is designed to compute each connection once. For this, the local and sparse tokens are selected such that there is no overlap between them during attention computation. Each query is connected to 3 local blocks and 2 sparse blocks of keys. The maximum context length (distance between two keys) is then equal to $3 \times b_t + 2 \times b_t \times f$. The concatenation of local and sparse keys is shown Figure \ref{fig:construction}. For causal attention, the third local block and the second sparse block can be ignored during computation.

\subsection{Global Attention}

Global tokens improve the flow of information inside the model. They attend to every tokens across the sequence and all tokens attend to them. Rather than picking a subset of tokens and defining them as global, they are prepended to the sequence and trained using their own embedding matrix, thus their number is an additional hyperparameter. When a model is converted to its LSG version, the first global token is initialized as the sum of the [CLS] (or \verb|<s>|) token and the first position from the positional embedding. The other global tokens are initialized as the sum of [MASK] (or \verb|<mask>|) token and the other positions from the positional embedding.

\subsection{Positional Embedding}
It is necessary to modify the positional embedding matrix to reuse existing models to process long sequences. 
Similarly to Longformer's authors \citep{Beltagy2020}, instead of randomly initializing the new positions, the original matrix is duplicated and concatenated until the desired max sequence length is reached.

\section{Experiments}

The LSG model is implemented in PyTorch and aims at performing model extrapolation by replacing full attention by the LSG attention in various architectures of the HuggingFace library. In the experiments, the official RoBERTa-base checkpoint for classification tasks and BART-base checkpoint for summarization tasks are extrapolated using LSG attention. All metrics are reported for the test set except in the case where only the validation set is available. We use a batch size of 32, a linear decaying learning rate, a dropout rate of 0.10 and Adam (0.9, 0.999) optimizer \citep{Kingma2014} for classification and summarization experiments.
An experiment comparing several attention approximations to extrapolate RoBERTa in an MLM task is first discussed as it is used to limit the number of tested alternatives, and therefore reduce the cost of the proposed evaluations. All experiments are run on NVIDIA Quadro RTX 8000 48Gb GPUs.

\subsection{RoBERTa extrapolation on MLM} \label{app:extrapolation}

A simple test on a MLM task is performed to test for the ability of an attention mechanism to extrapolate a model to longer sequences without additional training. To do so, a RoBERTa-base model is considered and two experiments are conducted. First, the full attention is substituted by different kinds of attention (kernel, factorization, local, fixed pattern) and each model is evaluated on sequences of the same length as those considered during RoBERTa initial training (512 tokens). For the second experiment, their ability to extrapolate to 4,096 tokens sequences without additional training is tested (the positional embedding being duplicated 8 times).

A random sample from Wikipedia + BookCorpus + CC\_News is used; BPC and MLM accuracy are reported in Table \ref{tab:results_mlm}. RoBERTa's author report a 1.880 BPC loss; we obtain a comparable loss of 1.881 on this random sample.
 
Only Longformer, Big Bird and LSG attention manage to obtain competing BPC while processing sequences of the same length as those considered during the original RoBERTa training. Other approaches such as Linformer, Performer or Reformer requires additional MLM fine-tuning to leverage an existing checkpoint. 
It can be seen that RoBERTa fails to extrapolate to longer sequences (+2.454 BPC), which highlights that full attention is not suitable for extrapolation. Longformer and Big Bird attention results show the capability of these approaches to perform some form of extrapolation. Therefore, we restrict our comparison to these two approaches in order to reduce the costs of our experimentation.

\begin{table*}[ht]

\centering
\begin{tabular}{l|cc|cc}
\multirow{2}{*}{\textbf{Attention}} & \multicolumn{2}{c|}{\textbf{512 length}} & \multicolumn{2}{c}{\textbf{4,096 length}} \\
 & BPC & Accuracy & BPC & Accuracy \\
\hline
RoBERTa  (full) \citep{Liu2019}                & 1.881 &  \textbf{0.732} & 4.335 & 0.359 \\
\hline
Linear Attn. \citep{Katharopoulos2020}  & 11.324 &  0.061 & 11.474 & 0.058 \\
Efficient Attn. \citep{Shen2020}        & 21.022 & 0.102 & 20.574 & 0.097 \\
Performer \citep{Choromanski2021}       & 10.382 & 0.107 & 10.556 & 0.102 \\
\hline
Linformer (128 proj.) \citep{Wang2020}        & 22.176 & 0.098 & 20.386 & 0.032 \\
Reformer \citep{Kitaev2020}             & 17.602 & 0.003 & 18.608 & 0.002 \\
\hline
Longformer (512) \citep{Beltagy2020}    & 1.929 & 0.726 & 2.051 & 0.708 \\
Big Bird (64) \citep{Zaheer2020}    & 1.881 & \textbf{0.732} & 2.439 & 0.659 \\
\hline
LSG-Norm (128/2) \quad \textbf{(block size / sparsity)}   & 1.919 & 0.727 & 2.032 & \textbf{0.712} \\
LSG-Stride (128/2)                         & 1.938 & 0.724 & 2.046 & 0.710 \\
LSG-BlockStride (128/2)                    & 1.940 & 0.724 & 2.048 & 0.709 \\
LSG-Pooling (128/2)                        & 1.968 & 0.720 & 2.064 & 0.706 \\
LSG-LSH (128/2)                            & 1.969 & 0.719 & 2.065 & 0.705 \\
\hline
\end{tabular}
\caption{BPC and MLM accuracy of RoBERTa-base with various Attention mechanisms.}
\label{tab:results_mlm}
\end{table*}

\subsection{Classification Tasks}
To evaluate the relevance of LSG, we compare our approach to common model architectures able to process long sequences with a similar number of parameters. Experiments are performed on Longformer \citep{Beltagy2020}, Big Bird \citep{Zaheer2020} and on all sparse attention types with a block size of 128 and a sparsify factor of 4. All models are fine-tuned on IMDb, ArXiv, Patent, Scotus, EcthrA and EcthrB datasets presented below.

\subsubsection{Datasets}
Datasets are available on the HuggingFace hub, see Appendix \ref{app:datasets}, e.g. detailed statistics in Table \ref{tab:datasets_sizes}.

\paragraph{IMDb } \citep{Maas2011} binary sentiment analysis classification task from movie reviews.

\paragraph{ArXiv} \citep{He2019} set of documents from ArXiv where the objective is to predict a topic from 11 available classes. Because there is no official split, a random one is made of 28K, 2.5K and 2.5K documents for train, validation and test.

\paragraph{Patent} \citep{Sharma2019} subset of the Big Patent summarization dataset. The task is redefined as a classification task where the objective is to predict the patent category using the full document (9 classes). A random split of 25K, 5K and 5K documents for train, validation and test is created.

\medskip

Some specific domains are highly dependent on processing long sequences, e.g. legal domain in which sentences tend to be long and complex. To demonstrate the ability of LSG attention to leverage pretrained models in such cases, the following three datasets are chosen from LexGlue \citep{Chalkidis2021}, a benchmark focused on legal documents. Tasks where the input is on average significantly longer than 512 tokens have been selected.

\paragraph{Scotus} Given a court opinion, the task is to predict the relevant issue area among 14 choices.

\paragraph{ECtHRa and ECtHRb} The objective is to predict which articles of the European Court of Human Rights (ECHR) have been violated (if any) from case description: multi-label task (10 + 1 labels).

\subsubsection{Training setup and architecture}

To make a fair comparison between models and architectures, fine-tuning is done with the same learning rate, number of steps (or epochs) and batch size. To show that the LSG attention is compatible with different architectures, the LexGlue tasks are also run with LEGAL-BERT \citep{Chalkidis2020} converted to its LSG version using the provided conversion tools.

\subsubsection{Results}
We report all experiment results in Table \ref{tab:classification}. We observe that LSG is competitive with Longformer and Big Bird models with input sequences up to 4096 tokens long. A major difference lies in the implementation itself since the LSG model is twice as fast to train on these lengths without additional memory cost; this aspect is discussed in Section \ref{characteristics}. 

On Patent, ECtHRa and ECtHRb tasks, the ability to process longer sequences improves significantly the F-measures compared to a vanilla (full attention) RoBERTa model. We also observe that Big Bird model is in general slightly under its counterpart except for the ECtHRb dataset. This probably comes from the random attention mechanism which may require additional training steps. LSG-LSH and Big Bird models are affected by randomness during inference, thus their performance can differ between runs. 

Extrapolating LEGAL-BERT with LSG to handle longer sequences improves predictions; this behavior is expected and has been observed by the authors of the LexGlue benchmark. The choice of the sparse attention is likely task specific. Using local attention only with a large block size is also a viable option. The role of global tokens is not discussed here since we only use one for all experiments. We show in the next section with summarization tasks the utility of such tokens.

\begin{table*}[ht]

\centering
\begin{tabular}{l|ccc|ccc}
\hline
 & \textbf{IMDb} & \textbf{Arxiv} & \textbf{Patent} & \textbf{Scotus} & \textbf{ECtHRa} & \textbf{ECtHRb}\\
 \hline
 Epochs                 & 3    & 3    & 3    & 7    & 5    & 5 \\
 Learning rate          & 2e-5 & 5e-5 & 2e-5 & 1e-4 & 1e-4 & 1e-4 \\
 \hline
 RoBERTa (512-length)          & 95.5 & 87.2/86.8 & 66.6/61.8 & 69.4/60.8 & 62.9/58.2 & 72.0/65.9 \\
 Longformer             & 95.9 & \textbf{88.2/87.9} & 69.8/63.8 & 72.9/62.6 & 68.3/59.7 & 78.9/72.2\\
 Big Bird ETC           & 95.4 & 85.9/85.5 & 69.4/63.9 & 69.4/58.2 & 68.3/60.3 & \textbf{80.0/70.6}\\
 \hline
 LSG-Local (256/0)        & \textbf{96.0} & 87.5/87.1 & 69.9/64.8 & \textbf{73.3/63.7} & 68.8/63.7 & 79.9/73.4 \\
 LSG-Stride (128/4)       & 95.6 & \textbf{88.2/87.9} & 69.2/64.0 & 70.5/60.0 & 69.5/62.3 & 79.3/71.6 \\
 LSG-BlockStride (128/4)  & 95.7 & 87.7/87.4 & 69.6/64.1 & 72.5/63.1 & 69.1/58.6 & 79.5/71.8 \\
 LSG-Norm (128/4)         & 95.7 & 87.0/86.6 & \textbf{70.0/64.4} & 71.3/60.8 & 70.1/61.9 & 79.4/72.1 \\
 LSG-Pooling (128/4)      & 95.9 & 87.5/87.3 & 69.4/64.1 & 72.6/60.9 & 70.2/61.4 & 79.0/73.1 \\
 LSG-LSH (128/4)          & 95.8 & \textbf{88.2/87.9} & 69.5/64.2 & 70.3/54.6 & \textbf{71.0/60.3} & 78.9/71.0\\
 \hline
 \hline
 Legal-BERT (512-length)  & - & - & - & 73.5/60.5 & 64.2/58.2 & 73.2/65.9 \\
 LSG-Legal-BERT (256/0)   & - & - & - & 74.5/62.6 & 71.7/63.9 & 81.0/75.1 \\
 \hline
\end{tabular}
\caption{\label{tab:classification}
Micro/Macro F-1 on classification datasets.
}
\end{table*}

\subsection{Summarization Tasks}
All summarization experiments are run using a 8e-5 learning rate, a 10\% warmup, a length penalty of 2.0 and a beam size of 5 for beam search. The validation set is used to choose the max generation length.
We choose to evaluate our models on summarization tasks where the input is significantly longer than 1k tokens only. We fine-tune our model on ArXiv, PubMed, MultiNews and MediaSum datasets we present below during respectively 6, 8, 12 and 6 epochs for 4,096-length inputs.

\subsubsection{Datasets}
The next datasets are available on the HuggingFace hub, see Appendix \ref{app:datasets}.

\paragraph{ArXiv and Pubmed} \citep{Cohan2018} are sets of documents from ArXiv and Pubmed; the goal is to generate an abstract using a document as input.

\paragraph{MultiNews} \citep{Alex2019} involves generating human-written summaries from sets of news documents.

\paragraph{MediaSum} \citep{Zhu2021} consists of using interview transcripts from CNN and NPR media to generate a summary.

We report detailed statistics in Table \ref{tab:datasets_sizes} in Appendix \ref{app:datasets}. The average length and the 90\% quantile of all documents and summaries using a whitespace separator are reported. Note that ArXiv abstracts are significantly longer in the training set (300 on average) than in the validation and test sets (173 on average). Most summarization models are limited to 1,000 tokens inputs, thus they are not able to process a full document to generate a summary. 

\subsubsection{Training setup and architecture}

We first convert the BART-base model \citep{Lewis2020} to its LSG version by replacing the full attention in the encoder part and adding global tokens. The model is then fine-tuned on 4,096-length inputs and evaluated. To reduce computational costs, experiments on 16,384-length inputs are warm started from the 4,096-length experiments using the conversion script. The model is then fine-tuned during a single epoch if necessary using the same training parameters. We propose 3 setups for the 16,384-length. First we evaluate the model with pure extrapolation from 4,096-length (no additional training). In the second setup, we extrapolate and add 64 global tokens we choose to fine-tune. In the last setup, we extrapolate, we add 64 global tokens and we fine-tune the full model. Extrapolation is done by concatenating 4 copies of the positional embedding matrix ($4 \times 4096$).

Compared to the existing literature, the model is rather small and an input sequence of 16384 tokens can fit on a 48Gb GPU during training without relying on gradient-checkpointing. The size of various summarization models from the literature are reported in Table \ref{tab:models_summarization}.

\begin{table}

\centering
\begin{tabular}{lc}
\hline
\textbf{Models} & Parameters \\
\hline
PRIMERA \citep{Xiao2022}        & 447M \\
LED \citep{Beltagy2020}         & 460M \\
HAT-BART \citep{Rohde2021}      & 471M \\
Pegasus \citep{Zhang2019}       & 577M \\
Big Bird-Peg. \citep{Zaheer2020} & 577M \\
Hepos \citep{Huang2021}         & 406M \\
\hline
LongT5-Base \citep{Guo2021}     & 220M \\ 
LongT5-L                        & 770M \\
LongT5-XL                       & 3B   \\
\hline
Ours, LSG-BART-base (256/0)     & 145M \\ 
\hline
\end{tabular}
\caption{\label{tab:models_summarization}
Parameters count of summarization models.}
\end{table}

\subsubsection{Results}
LSG-BART is compared to state-of-the-art models by reporting the results from their respective papers. We use ROUGE-1, ROUGE-2 and ROUGE-L evaluation metrics as comparison points. 

As shown in Tables \ref{tab:arxiv_summarization}, \ref{tab:pubmed_summarization}, \ref{tab:multinews_summarization} and \ref{tab:mediasum_summarization}, our approach can achieve competitive performances with a limited size without pretraining a new model from scratch. The second important element is the ability of this approach to improve metrics from 4.096 to 16.384-length inputs without additional fine-tuning, this is especially true on ArXiv and PubMed datasets which have the longest input sequences. Fine tuning additional global tokens further improveS metrics while limiting cost and training time compared to a fully tuned model.

On the ArXiv dataset (Table \ref{tab:arxiv_summarization}), a max sequence generation of 320 tokens is chosen, our approach is competitive with every size of the LongT5 model. However, the authors pointed out that they used greedy generation instead of beam search, thus their results are likely underestimated.  

\begin{table}

\centering
\begin{tabular}{lccc}
\hline
\textbf{Models} & \textbf{R1} & \textbf{R2} & \textbf{RL} \\
\hline
Pegasus (1K)           & 44.70 & 17.27 & 25.80 \\
Big Bird-Peg. (4K)      & 46.63 & 19.02 & 41.77 \\
\hline
LED (4K)               & 44.40 & 17.94 & 39.76 \\
LED (16K)              & 46.63 & 19.62 & 41.83 \\ 
PRIMERA (4K)           & 47.58 & 20.75 & 42.57 \\
\hline 
HAT-BART (4K)          & 46.68 & 19.07 & 42.17 \\
\hline
Hepos-LSH (7.2K)       & 48.24 & 20.26 & 41.78 \\
Hepos-SKN (10.2K)      & 47.87 & 20.00 & 41.50 \\
\hline
LongT5-Base (4K)       & 44.87 & 18.54 & 40.97 \\
LongT5-L (16K)         & 48.28 & 21.63 & 44.11 \\
LongT5-XL (16K)        & 48.35 & 21.92 & 44.27 \\
\hline
\hline
Ours (4K)              & 46.65 & 18.91 & 42.18 \\ 
Ours (16K)             & 47.03 & 20.19 & 42.69 \\ 
\quad + global tuning  & 48.08 & 20.42 & 43.65 \\ 
\quad + full tuning    & 48.74 & 20.88 & 44.23 \\
\hline
\end{tabular}
\caption{\label{tab:arxiv_summarization}
ROUGE performances on ArXiv dataset.}
\end{table}

On the PubMed dataset (Table \ref{tab:pubmed_summarization}), a max sequence generation of 512 tokens is chosen, our approach is close to Hepos models which also rely on BART. LongT5 is significantly better here and this difference may be related to the way this model is pretrained and the dataset used for this.

\begin{table}

\centering
\begin{tabular}{lccc}
\hline
\textbf{Models} & \textbf{R1} & \textbf{R2} & \textbf{RL} \\
\hline
Pegasus (1K)           & 45.49 & 19.90 & 27.69 \\
Big Bird-Peg. (4K)      & 46.32 & 20.65 & 42.33 \\
\hline
HAT-BART (4K)          & 48.36 & 21.43 & 37.00 \\
\hline 
Hepos-LSH (7.2K)       & 48.12 & 21.06 & 42.72 \\
Hepos-SKN (10.2K)      & 47.93 & 20.74 & 42.58 \\
\hline
LongT5-Base (4K)       & 47.77 & 22.58 & 44.38 \\
LongT5-L (16K)         & 49.98 & 24.69 & 46.46 \\
LongT5-XL (16K)        & 50.23 & 24.76 & 46.67 \\
\hline
\hline
Ours (4K)              & 47.37 & 21.74 & 43.67 \\ 
Ours (16K)             & 48.03 & 22.42 & 44.32 \\ 
\quad + global tuning  & 48.12 & 20.46 & 44.40 \\ 
\quad + full tuning    & 48.32 & 22.52 & 44.57 \\
\hline
\end{tabular}
\caption{\label{tab:pubmed_summarization}
ROUGE performances on PubMed dataset.}
\end{table}

On the MultiNews dataset (Table \ref{tab:multinews_summarization}), a max sequence generation of 320 tokens is chosen, our approach is close again to the LongT5 models. While extrapolation improves metrics, additional fine-tuning has a negative impact. Since this dataset is rather small (45K examples, ~1,400 steps), fine-tuning a single epoch is not enough for the model to converge properly, longer training is required.

\begin{table}

\centering
\begin{tabular}{lccc}
\hline
\textbf{Models} & \textbf{R1} & \textbf{R2} & \textbf{RL} \\
\hline
TG-MultiSum            & 47.10 & 17.55 & 20.73 \\
PRIMERA (4K)           & 49.90 & 21.10 & 25.9  \\
\hline
LongT5-Base (4K)       & 46.01 & 17.37 & 23.50 \\
LongT5-L (4K)          & 46.99 & 18.21 & 24.08 \\
LongT5-L (8K)          & 47.18 & 18.44 & 24.18 \\
LongT5-XL (8K)         & 48.17 & 19.43 & 24.90 \\
\hline
\hline
Ours (4K)              & 47.10 & 18.94 & 25.22 \\ 
Ours (16K)             & 47.30 & 19.19 & 25.38 \\ 
\quad + global tuning  & 47.23 & 19.18 & 25.29 \\ 
\quad + full tuning    & 47.07 & 19.04 & 25.35 \\
\hline
\end{tabular}
\caption{\label{tab:multinews_summarization}
ROUGE performances on MultiNews.}
\end{table}

On the MediaSum dataset (Table \ref{tab:mediasum_summarization}), a max sequence generation of 128 tokens is chosen. Our approach is close to the LongT5-base model again. This dataset has the shortest inputs, thus processing a maximum of 16,384 tokens has a marginal impact on performances. 

Additional results using different types of sparse attention are detailed in Appendix \ref{app:summarization}.

\begin{table}

\centering
\begin{tabular}{lccc}
\hline
\textbf{Models} & \textbf{R1} & \textbf{R2} & \textbf{RL} \\
\hline
BART-Large (1K)        & 35.09 & 18.05 & 31.44 \\
T5-large (1K)          & 30.68 & 14.88 & 27.88 \\
\hline
LongT5-Base (4K)       & 35.09 & 18.35 & 31.87 \\
LongT5-L (4K)          & 35.54 & 19.04 & 32.20 \\
LongT5-XL (4K)         & 36.15 & 19.66 & 32.80 \\
\hline
\hline
Ours (4K)              & 35.16 & 18.13 & 32.20 \\ 
Ours (16K)             & 35.17 & 18.13 & 32.21 \\ 
\quad + global tuning  & 35.22 & 18.08 & 32.22 \\ 
\quad + full tuning    & 35.31 & 18.35 & 32.47 \\
\hline
\end{tabular}
\caption{\label{tab:mediasum_summarization}
ROUGE performances on MediaSum.}
\end{table}




\begin{table}

\centering
\begin{tabular}{lcc}
\hline
\textbf{Models} & \textbf{Time/step} & \textbf{Memory} \\
\hline
RoBERTa (512)   & 1.18 s & 28.8/32.1 Gb \\
\hline
Longformer      & 3.27 s & 39.2/38.1 Gb \\
Big Bird         & 2.89 s & 44.5/44.4 Gb \\
\hline
LSG-Local (256/0)       & 1.42 s & 40.7/32.3 Gb \\
LSG-Norm (128/4)  & 1.52 s & 40.4/33.4 Gb \\
LSG-Norm (32/4)        & 1.24 s & 26.1/24.3 Gb \\
\hline
\end{tabular}
\caption{\label{tab:model_speed}
Training speed and memory with a batch of 16384 tokens (Adam optimizer). All models rely on sequences of 4096 tokens except RoBERTa. Memory usage is computed with and without attention dropout.}
\end{table}

\section{Implementation details}\label{characteristics}

The proposed implementations are exclusively based on those of HuggingFace in which the global tokens are prepended to the sequence and the attention layer is replaced by its efficient version; other elements are not modified.

To improve efficiency, the inputs are split into blocks. Each block of queries is connected to 3 blocks of local keys, 2 blocks of sparse keys and to all global keys. Thus for head $h$, queries, keys and values are of shape $\bm{Q}^h \in \mathbb{R}^{n_b \times b_t \times d_h}$ and $\bm{K}^h, \bm{V}^h \in \mathbb{R}^{n_b \times (5b_t + g) \times d_h}$ with $n_b$ the number of blocks, $b_t$ the size of blocks, $d_h$ the size of the head and $g$ the number of global tokens. This format improves computational speed as shown in Table \ref{tab:model_speed}. Global attention is computed independently.

\section{Conclusion}
We have presented LSG attention, a novel efficient $O(n)$ alternative to the full attention mechanism relying on local, sparse and global attentions. Our results on MLM, classification and summarization tasks show that LSG is a competitive full attention substitute for pretrained Transformers to efficiently extrapolate to long input sequences. 
We also proposed an optimized implementation of the LSG attention mechanism on HuggingFace, improving training speed by a factor of 2 without additional memory cost compared to Longformer and Big Bird models.
By providing a conversion tool to leverage existing models and checkpoints (BERT, RoBERTa, DistilBERT, BART), the proposed approach removes the need of a costly re-training of existing models to handle long sequences. 

\section*{Limitations}
Although the proposed conversion tool allows to convert existing checkpoints of commonly used models, it is today necessary to reimplement the approach for each architecture due to the lack of homogeneity of HuggingFace implementations (no wrapper available yet). Maintenance may therefore be a problem in the long run to ensure compatibility. We however provide implementation examples as well as documentation to ease the conversion process.

Concerning the proposed attention itself, the choice of the sparse attention remains an additional hyperparameter which is task specific. There is no rule of thumb to choose the sparse type, the size of blocks and the sparsity factor. The role of global tokens is also debatable. Their use can slow down training speed and convergence. In practice, the impact of these tokens is positive if the model is trained a sufficient number of steps.

Although the approach allows an existing model to be reused without having to pretrain from scratch and to reduce the duration of fine-tuning phases, the complexity remains linear with the length of the input. This does not eliminate the energy costs required to deploy Transformer models.

\section*{Acknowledgements}
This work has benefited from LAWBOT (ANR-20-CE38-0013) grant and HPC resources of IDRIS (allocation 2022-AD011011309R2) made by GENCI.
\bibliography{main}
\bibliographystyle{unsrt}

\appendix

\section{Training parameters}

We use a batch size of 32, a linear decaying learning rate, a dropout rate of 0.10 and Adam optimizer for all tasks. Other parameters are reported in Table \ref{tab:training_params}.

\begin{table}[ht]

\centering
\begin{tabular}{lccc}
\hline
 & \textbf{Epochs} & \textbf{LR} & \textbf{Warmup}\\
\hline
\textbf{Classification} & \\
\hline
IMDb & 3 & 2e-5 & 0\% \\
ArXiv & 3 & 5e-5 & 0\% \\
Patent & 3 & 2e-5 & 0\% \\
Scotus & 7 & 1e-4 & 0\% \\
ECtHRa & 5 & 1e-4 & 0\% \\
ECtHRb & 5 & 1e-4 & 0\% \\
\hline
\textbf{Summarization} & \\ 
\hline
ArXiv & 6/1 & 8e-5 & 10\% \\
PubMed & 8/1 & 8e-5 & 10\% \\
MultiNews & 12/1 & 8e-5 & 10\% \\
MediaSum & 6/1 & 8e-5 & 10\% \\
\hline
\end{tabular}
\caption{\label{tab:training_params}
Training parameters for all tasks.}
\end{table}

\section{Additional classification results} \label{app:classification}

Additional classification results using a smaller block size are presented in Table \ref{tab:appendix_classification}. It shows that the use of a smaller block size remains competitive even if a slight loss in performance is observed. 

\begin{table*}[ht!]

\centering
\begin{tabular}{l|ccc|ccc}
\hline
 & \textbf{IMDb} & \textbf{Arxiv} & \textbf{Patent} & \textbf{Scotus} & \textbf{ECtHRa} & \textbf{ECtHRb}\\
 \hline
 RoBERTa (512-length)          & 95.5 & 87.2/86.8 & 66.6/61.8 & 69.4/60.8 & 62.9/58.2 & 72.0/65.9 \\
 Longformer             & 95.9 & \textbf{88.2/87.9} & 69.8/63.8 & 72.9/62.6 & 68.3/59.7 & 78.9/72.2\\
 Big Bird            & 95.4 & 85.9/85.5 & 69.4/63.9 & 69.4/58.2 & 68.3/60.3 & 80.0/70.6\\
 \hline
 LSG-Local (256/0)        & \textbf{96.0} & 87.5/87.1 & 69.9/64.8 & \textbf{73.3/63.7} & 68.8/63.7 & 79.9/73.4 \\
 LSG-Stride (128/4)       & 95.6 & \textbf{88.2/87.9} & 69.2/64.0 & 70.5/60.0 & 69.5/62.3 & 79.3/71.6 \\
 LSG-BlockStride (128/4)  & 95.7 & 87.7/87.4 & 69.6/64.1 & 72.5/63.1 & 69.1/58.6 & 79.5/71.8 \\
 LSG-Norm (128/4)         & 95.7 & 87.0/86.6 & \textbf{70.0/64.4} & 71.3/60.8 & 70.1/61.9 & 79.4/72.1 \\
 LSG-Pooling (128/4)      & 95.9 & 87.5/87.3 & 69.4/64.1 & 72.6/60.9 & 70.2/61.4 & 79.0/73.1 \\
 LSG-LSH (128/4)          & 95.8 & \textbf{88.2/87.9} & 69.5/64.2 & 70.3/54.6 & \textbf{71.0/60.3} & 78.9/71.0\\
 \hline
 LSG-Stride (32/4)       & 95.6 & 85.0/84.5 & 69.2/63.0 & 72.4/62.7 & 69.9/58.4 & 79.2/72.3 \\
 LSG-BlockStride (32/4)  & 95.4 & 86.6/86.3 & 69.4/63.4 & 72.4/62.6 & 70.3/60.9 & 79.2/69.4 \\
 LSG-Norm (32/4)         & 95.7 & 85.3/84.9 & 69.3/63.9 & 72.2/62.5 & 68.9/60.9 & 79.1/73.5 \\
 LSG-Pooling (32/4)      & 95.7 & 88.2/88.0 & 69.2/63.3 & 72.6/60.2 & 69.6/59.1 & 79.3/72.1 \\
 LSG-LSH (32/4)          & 95.7 & 88.0/87.7 & 68.9/62.9 & 71.9/60.4 & 70.0/61.1 & \textbf{80.3/73.1} \\
 \hline
 \hline
 Legal-BERT (512-length)       & - & - & - & 73.5/60.5 & 64.2/58.2 & 73.2/65.9 \\
 LSG-Legal-BERT        & - & - & - & 74.5/62.6 & 71.7/63.9 & 81.0/75.1 \\
 \hline
\end{tabular}
\caption{\label{tab:appendix_classification}
Micro/Macro F-1 on classification datasets.
}
\end{table*}

\section{Additional summarization results} \label{app:summarization}

\begin{table}[H]

\centering
\begin{tabular}{lccc|c}
\hline
\textbf{Models (4,096)} & \textbf{R1} & \textbf{R2} & \textbf{RL} & \textbf{Ctx.} \\
\hline
\textbf{32/4} & & & \\
\hline
Pooling                & 42.75 & 16.34 & 38.23 & 160 \\ 
Stride 	               & 44.23 & 17.21 & 39.72 & 160 \\ 
Block Stride           & 44.15 & 17.10 & 39.60 & 160 \\ 
Norm                   & 42.02 & 15.65 & 37.45 & 160 \\ 
LSH                    & 42.58 & 16.21 & 38.04 & 160 \\ 
\hline
\textbf{128/4} & & & \\
\hline
Pooling                & 46.27 & 18.68 & 41.82 & 644 \\ 
Stride 	               & 46.34 & 18.64 & 41.87 & 644 \\ 
Block Stride           & 46.23 & 18.62 & 41.80 & 644 \\ 
Norm                   & 45.96 & 18.46 & 41.51 & 644 \\ 
LSH                    & 46.19 & 18.72 & 41.76 & 644 \\ 
\hline
\hline
Reference              & 46.65 & 18.91 & 42.18 & 784 \\ 
\hline
\end{tabular}
\caption{\label{tab:arxiv_summarization_substitution}
ROUGE performances on ArXiv dataset.}
\end{table}

\begin{table}[H]

\centering
\begin{tabular}{lccc|c}
\hline
\textbf{Models (4,096)} & \textbf{R1} & \textbf{R2} & \textbf{RL} & \textbf{Ctx.} \\
\hline
\textbf{32/4} & & & \\
\hline
Pooling                & 44.60 & 19.35 & 40.85 & 160 \\ 
Stride 	               & 45.52 & 20.07 & 41.75 & 160 \\ 
Block Stride           & 45.30 & 19.89 & 41.54 & 160 \\ 
Norm                   & 44.30 & 19.05 & 40.47 & 160 \\ 
LSH                    & 44.53 & 19.27 & 40.74 & 160 \\ 
\hline
\textbf{128/4} & & & \\
\hline
Pooling                & 47.11 & 21.42 & 43.40 & 644 \\ 
Stride 	               & 47.16 & 21.49 & 43.44 & 644 \\ 
Block Stride           & 47.13 & 21.46 & 43.42 & 644 \\ 
Norm                   & 47.09 & 21.44 & 43.36 & 644 \\ 
LSH                    & 47.11 & 21.41 & 43.42 & 644 \\ 
\hline
\hline
Reference              & 47.37 & 21.74 & 43.67 & 784 \\ 
\hline
\end{tabular}
\caption{\label{tab:pubmed_summarization_substitution}
ROUGE performances on PubMed dataset.}
\end{table}

Trained models on summarization tasks are reevaluated after changing the type of sparse attention and block size. Results on ArXiv and PubMed are reported in Tables \ref{tab:arxiv_summarization_substitution} and \ref{tab:pubmed_summarization_substitution}. The context column refers to the number of keys each query attends to. As reference models are trained on large (256) local blocks, the number of connections is $3 \times 256$. By using 20\% less keys (644), inference results are still competitive even though the model has never seen these specific sparse patterns before. By limiting connections to 20\% of the keys (160), a performance drop is observed even if the metrics still remain respectable. Under these conditions, stride and block-stride approaches generate better predictions.

\section{Datasets and Models} \label{app:datasets}

All the datasets evalued are available on the HuggingFace hub, links are provided in Table \ref{tab:datasets_links}.

\begin{table}[ht]

\centering
\begin{tabular}{lc}
\hline
 & \textbf{Link to the hub} \\
\hline
\textbf{Classification} & \\
\hline
IMDb & \href{https://huggingface.co/datasets/imdb}{imdb}\\
ArXiv & \href{https://huggingface.co/datasets/ccdv/arxiv-classification}{arxiv-classification} \\
Patent & \href{https://huggingface.co/datasets/ccdv/patent-classification}{patent-classification} \\
Scotus & \href{https://huggingface.co/datasets/lex_glue}{lex\_glue/scotus} \\
ECtHRa & \href{https://huggingface.co/datasets/lex_glue}{lex\_glue/ecthr\_a} \\
ECtHRb & \href{https://huggingface.co/datasets/lex_glue}{lex\_glue/ecthr\_b} \\
\hline
\textbf{Summarization} & \\ 
\hline
ArXiv & \href{https://huggingface.co/datasets/scientific_papers}{scientific\_papers/arxiv} \\
PubMed & \href{https://huggingface.co/datasets/scientific_papers}{scientific\_papers/pubmed} \\
MultiNews & \href{https://huggingface.co/datasets/multi_news}{multi\_news} \\
MediaSum & \href{https://huggingface.co/datasets/ccdv/mediasum}{mediasum} \\
\hline
\end{tabular}
\caption{\label{tab:datasets_links}
Links to datasets.}
\end{table}

Summarization checkpoints are available on the HuggingFace hub, links are provided in Table \ref{tab:models_links}. 

\begin{table}[ht]

\centering
\begin{tabular}{lc}
\hline
\textbf{Summarization} & \textbf{Link to the hub} \\
\hline
 LSG-PubMed (4,096)   & \href{https://huggingface.co/ccdv/lsg-bart-base-4096-pubmed}{lsg-pubmed} \\
 LSG-PubMed (16,384)  & \href{https://huggingface.co/ccdv/lsg-bart-base-16384-pubmed}{lsg-pubmed} \\
 \hline
 LSG-ArXiv (4,096)   & \href{https://huggingface.co/ccdv/lsg-bart-base-4096-arxiv}{lsg-arxiv} \\
 LSG-ArXiv (16,384)  & \href{https://huggingface.co/ccdv/lsg-bart-base-16384-arxiv}{lsg-arxiv} \\
 \hline
 LSG-MediaSum (4,096)   & \href{https://huggingface.co/ccdv/lsg-bart-base-4096-mediasum}{lsg-mediasum} \\
 LSG-MediaSum (16,384)   & \href{https://huggingface.co/ccdv/lsg-bart-base-16384-mediasum}{lsg-mediasum} \\
 \hline
 LSG-MultiNews (4,096)   & \href{https://huggingface.co/ccdv/lsg-bart-base-4096-multinews}{lsg-multinews} \\
 \hline
  \hline
 Additional checkpoints   & \href{https://huggingface.co/ccdv}{lsg-checkpoints} \\
\hline
\end{tabular}
\caption{\label{tab:models_links}
Links to model checkpoints.}
\end{table}

\begin{table*}[ht!]

\centering
\begin{tabular}{l|ccc|cc|cc}
\hline
\multirow{2}{*}{\textbf{Datasets}} & \multicolumn{3}{c|}{\textbf{Example count}} & \multicolumn{2}{c|}{\textbf{Input length}} & \multicolumn{2}{c}{\textbf{Summary length}}\\
                  & Train   & Validation & Test    & Average & q-90\% & Average & q-90\% \\
\hline
\textbf{Classification} \\ 
\hline
IMDb                & 25,000 & - & 25,000 & 234 & 458 & - & - \\
ArXiv               & 28,000 & 2,500 & 2,500 & 9,115 & 16,264 & - & - \\
Patent              & 25,000 & 5,000 & 5,000 & 3,593 & 6,579 & - & - \\
Scotus              & 5,000	 & 1,400 & 1,400 & 6,992 & 14,570 & - & - \\
ECtHRa              & 9,000	 & 1,000 & 1,000 & 1,681 & 3,728 & - & - \\
ECtHRb              & 9,000	 & 1,000 & 1,000 & 1,681 & 3,728 & - & - \\
\hline
\textbf{Summarization} \\ 
\hline
ArXiv             & 203,037 & 6,436      & 6,440   & 6,030   & 11,165 & 272     & 321 \\
PubMed            & 119,924 & 6,633      & 6,658   & 3,050   & 5,737  & 202     & 304 \\
MultiNews         & 44,972 & 5,622       & 5,622   & 1,792   & 3,375  & 217     & 294 \\
MediaSum          & 443,596 & 10,000     & 10,000  & 1,582   & 2,883  & 15      & 32  \\
\hline
\end{tabular}
\caption{\label{tab:datasets_sizes}
Statistics of evaluated datasets, lengths are whitespace based.}
\end{table*}

Average input size are provided in Table \ref{tab:datasets_sizes}. Note that token counts are obtained using a whitespace split. Subword tokenization increases these numbers by 30\% to 40\% depending on the tokenizer and the vocabulary size. 

\section{Hyperparameters and complexity} \label{app:complexity}

LSG attention is sensitive to several hyperparameters and the nature of the pretrained model.

\paragraph{Block size} Generally speaking, block size improves performance to a certain extent. If the converted model is trained on 512-length sequences, a block size beyond 256 has a negative impact and requires a longer fine-tuning phase. A smaller block reduces training and memory cost.

\paragraph{Sparsity factor} The sparsity factor is generally chosen between 0 (no sparse attention), 2, 4 and 8. Although the choice remains task-dependent, a factor above 8 tends on average to decrease the level of performance, especially for pooling-based approaches. This hyperparameter is chosen to be small when the task focuses on local information (MLM, NER) and can be larger for tasks requiring a wider context (summarization, question-answering).

\paragraph{Global tokens} The more global tokens there are, the longer it takes for the model to converge and obtain performance gains. The initialization of these tokens is important, in particular for the first one which is used as a pooling token for classification tasks ([CLS] or \verb|<s>| + position 0).

\begin{figure*}[ht!]
\includegraphics[width=1\linewidth]{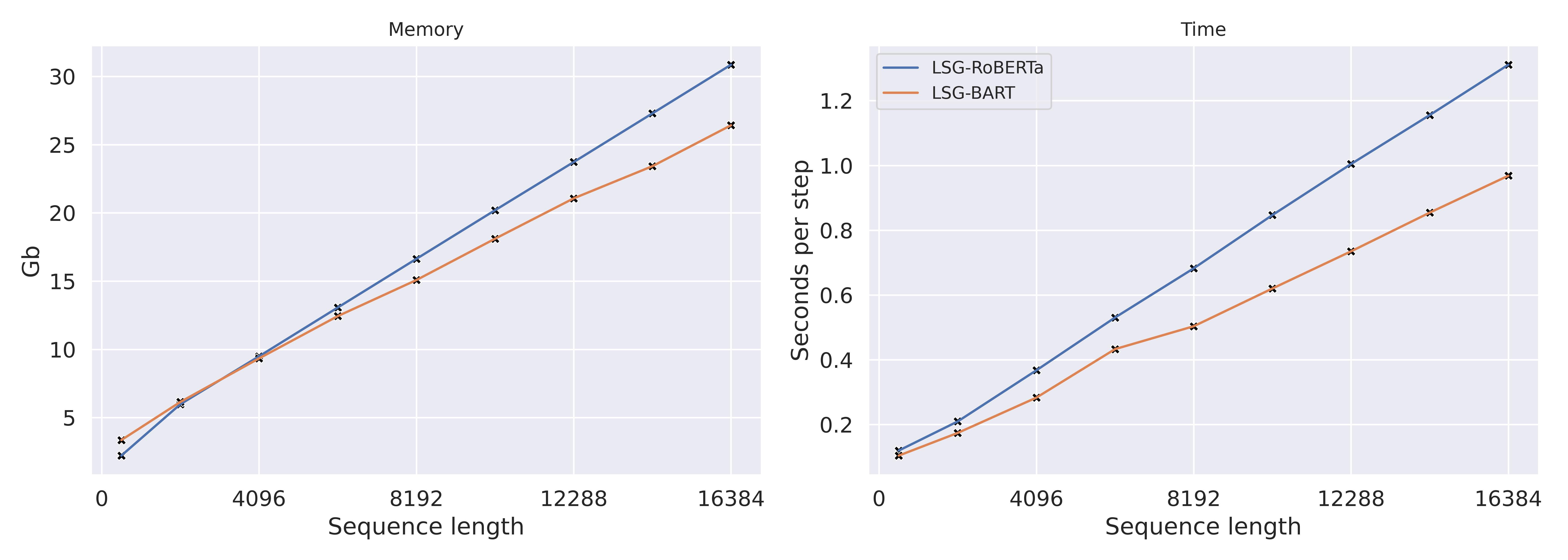}
\caption{\label{fig:complexity}
Memory and time complexity as a function of input sequence length.}
\end{figure*}

\paragraph{Overall complexity} LSG attention has some similarities with Big Bird attention and has the same $O(n)$ complexity with regard to sequence length. Figure \ref{fig:complexity} shows the effect of increasing sequence length on training time and memory consumption (with Adam optimizer).

\end{document}